\documentclass[runningheads]{llncs}

\usepackage{eccv}

\usepackage{eccvabbrv}

\usepackage{graphicx}
\usepackage{booktabs}

\usepackage[accsupp]{axessibility}  

\usepackage[pagebackref,breaklinks,colorlinks,citecolor=eccvblue]{hyperref}

\usepackage{orcidlink}

\usepackage{pifont}

\usepackage{diagbox}
\usepackage{graphicx}
\usepackage{amsmath}
\usepackage{amssymb}
\usepackage{booktabs}
\usepackage{multirow}
\usepackage{makecell}
\usepackage{float}
\usepackage{blindtext}
\usepackage{multirow}
\usepackage{bm}
\usepackage{bbm} 
\usepackage{comment}
\usepackage{algorithm}
\usepackage{algorithmic}
\usepackage{array} 
\usepackage{wrapfig}
\usepackage[dvipsnames]{xcolor}
\usepackage[safe]{tipa} 
\usepackage{color, colortbl}
\usepackage[misc,geometry]{ifsym} 
\makeatletter
\renewcommand\paragraph{
  \@startsection{paragraph} 
  {4} 
  {\z@} 
  {.5em \@plus1ex \@minus.2ex} 
  {-1.5em} 
  {\normalfont\normalsize\bfseries} 
}
\makeatother

\makeatletter
\def\@fnsymbol#1{\ensuremath{\ifcase#1\or \textsuperscript{~\Letter}\or \ddagger\or
   \mathsection\or \mathparagraph\or \|\or **\or \dagger\dagger
   \or \ddagger\ddagger \else\@ctrerr\fi}}
\makeatother

\makeatletter
\newcommand{\myfnsymbol}[1]{%
  \expandafter\@myfnsymbol\csname c@#1\endcsname
}
\newcommand{\@myfnsymbol}[1]{%
  \ifcase #1
  \or 1
  \or 2
  \or \TextOrMath{\textasteriskcentered}{*}
  \or \TextOrMath{\textdagger}{\dagger}
  \fi
}
\newcommand{\affiliationA}{\@myfnsymbol{1}}
\newcommand{\affiliationB}{\@myfnsymbol{2}}
\newcommand{\equalcontributor}{\@myfnsymbol{3}}
\newcommand{\correspondingA}{\@myfnsymbol{4}}
\makeatother

\definecolor{tabhighlight}{HTML}{e5e5e5}
\definecolor{citecolor}{HTML}{0071bc}
\begin{document}

\title{Quantized Prompt for Efficient Generalization of Vision-Language Models} 


\author{Tianxiang Hao\inst{1,2}\and
Xiaohan Ding\inst{3}\textsuperscript{(\Letter)}\and
Juexiao Feng\inst{1,2}\and
Yuhong Yang\inst{1,2}\and\\
Hui Chen\inst{2}\and
Guiguang Ding\inst{1,2}\textsuperscript{(\Letter)}
}

\authorrunning{T. Hao et al.}

\institute{School of Software, Tsinghua University, Beijing, China\and
BNRist, Beijing, China\and
Bytedance\\
\email{\{beyondhtx,xiaohding,suise.con,jichenhui\}@gmail.com\\fjx20@mails.tsinghua.edu.cn,\quad dinggg@tsinghua.edu.cn}\\}

\maketitle

\begin{abstract}
  In the past few years, large-scale pre-trained vision-language models like CLIP have achieved tremendous success in various fields. Naturally, how to transfer the rich knowledge in such huge pre-trained models to downstream tasks and datasets becomes a hot topic. During downstream adaptation, the most challenging problems are overfitting and catastrophic forgetting, which can cause the model to overly focus on the current data and lose more crucial domain-general knowledge. Existing works use classic regularization techniques to solve the problems. As solutions become increasingly complex, the ever-growing storage and inference costs are also a significant problem that urgently needs to be addressed. While in this paper, we start from an observation that proper random noise can suppress overfitting and catastrophic forgetting. Then we regard quantization error as a kind of noise, and explore quantization for regularizing vision-language model, which is quite efficiency and effective. Furthermore, to improve the model's generalization capability while maintaining its specialization capacity at minimal cost, we deeply analyze the characteristics of the weight distribution in prompts, conclude several principles for quantization module design and follow such principles to create several competitive baselines. The proposed method is significantly efficient due to its inherent lightweight nature, making it possible to adapt on extremely resource-limited devices. Our method can be fruitfully integrated into many existing approaches like MaPLe, enhancing accuracy while reducing storage overhead, making it more powerful yet versatile. Extensive experiments on 11 datasets shows great superiority of our method sufficiently. Code is available at \href{https://github.com/beyondhtx/QPrompt}{github}
\end{abstract}

\section{Introduction}
\label{sec:intro}
Recently, deep learning models and related technologies have seen rapid development~\cite{he2016deep,lyu2023box,lyu2024one,xiong2024temporal,xiong2023confidence,dosovitskiy2020image,houlsby2019parameter,hu2021lora,ding2021repvgg,ding2022scaling,ding2019acnet}. Vision-language model (VLM) like CLIP~\cite{radford2021learning} is one of the hottest research topics and leads to huge success. The excellent generalization ability is a crucial cornerstone of such achievements \cite{radford2021learning,jia2021align,yao2022filip,rao2022denseclip,mu2022slip}. 

When people have access to downstream data, it is better to tune the pre-trained VLM on the target dataset for higher accuracy. However, such full fine-tuning could easily cause the model to overfit the small downstream dataset and face catastrophic forgetting problem, leading to severe performance drop. 

To solve the problem, in this paper we will start from rethinking the relationship between noise and generalization. We propose to think noise as a kind of regularization techniques, which may be helpful for alleviating overfitting and catastrophic forgetting problem. As in \cref{subsec:rethinking_noise} and \cref{fig:motivation}, we then find that directly adding some random Gaussian noise to the tunable prompts of vision-language models would result in a performance gain in several cases, which partly validates our conjecture about using noise as a form of regularization. In particular, excessive noise diminishes the model's adaptation capability, while insufficient noise fails to provide effective regularization. Only noise of moderate intensity is beneficial for the model's generalization.

However, Gaussian noise is absolutely random and hard to control, and thus it is difficult for us to take full advantage of such type of noise to benefit generalization. Instead, we point out that quantization error is also a form of noise, and therefore, it is also possible to leverage quantization error to enhance the model's generalization performance. With this idea in mind, we thoroughly analyzed the distribution pattern of prompt weights in \cref{subsec:characteristics} and derived several design principles for quantization algorithms based on the observed phenomena. In \cref{sec:promptquant}, following the principles we summarized, we successfully designed an efficient quantization-aware training algorithm, which largely enhances the model's generalization ability while quantizing.


\begin{figure}[tb]
  \centering
    \includegraphics[width=0.6\linewidth]{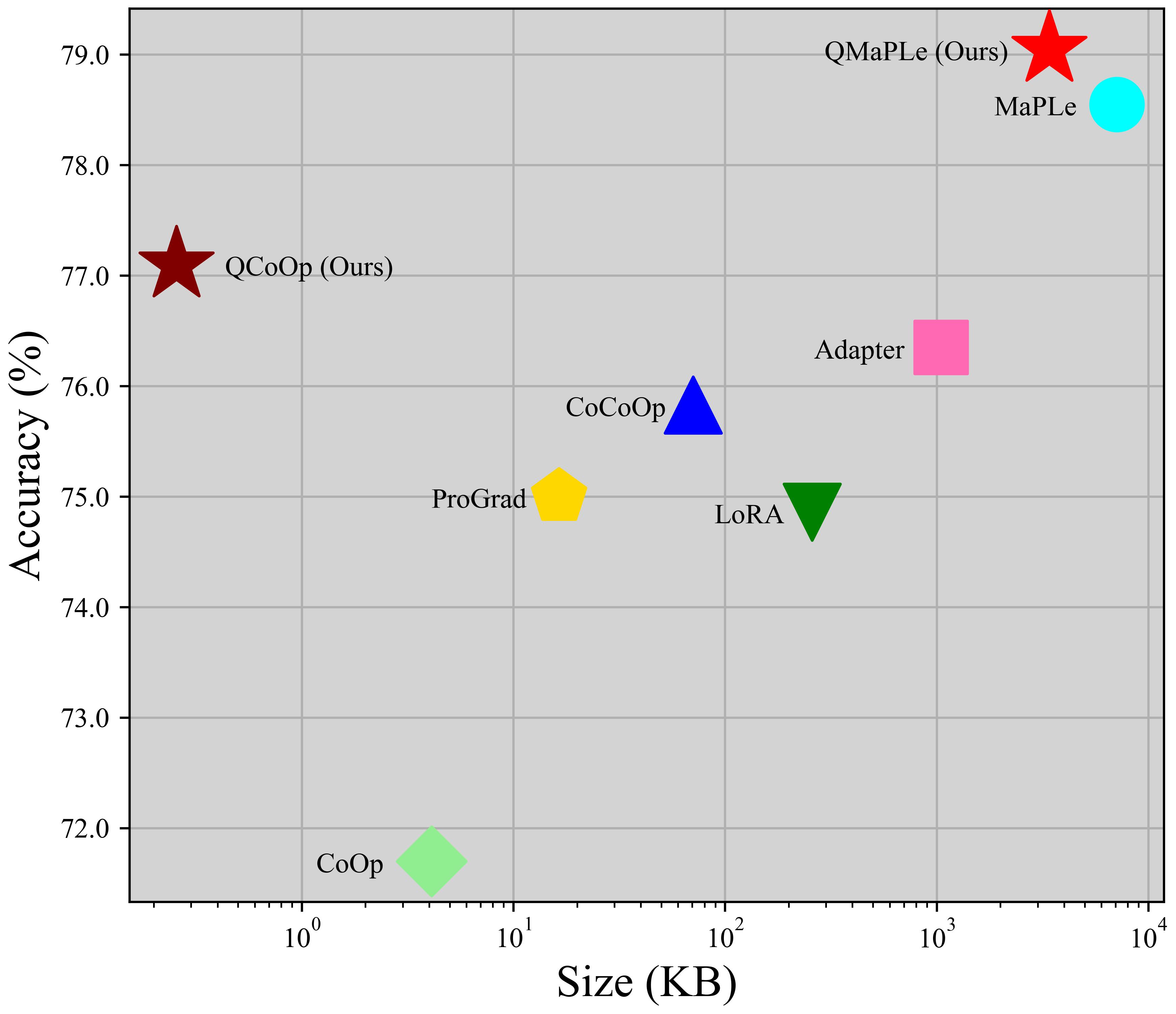}
  \setlength{\abovecaptionskip}{-0.01cm}
  \setlength{\belowcaptionskip}{-0.3cm}
  \caption{Overall performance comparison with existing vision-language tuning methods. Our method outperforms all of the state-of-the-art competitors with significantly fewer storage space. Based on the proposed quantization algorithm, our method could be integrated into many of the existing methods and bring consistent improvements with excellent efficiency.}
  \label{fig:performance}
\end{figure}



Since our design is general, we could fruitfully integrate our quantization strategy into many existing methods and reach a higher accuracy with much smaller model size due to the lightweight nature of quantization. For example, we integrate our method onto an existing popular method MaPLe, and our QMaPLe earns 0.57\% accuracy improvement with only $0.25\times$ size. 

In conclusion, we summarize our contribution as follows:

\vspace{-1mm}
\begin{itemize}
\setlength{\topsep}{0pt}
\setlength{\parsep}{0pt}
\setlength{\itemsep}{0pt}
\setlength{\partopsep}{0pt}
\setlength{\parskip}{0pt}
    \item We deeply analyze the effect of noise and rethink the relationship between noise and generalization in \cref{subsec:rethinking_noise} for vision-language models. As a result, we confirm that moderate noise would promote the model generalization
    \item We are the first to propose to quantize prompts. By detailed observation and hard thinking, we conclude several principles about how to effectively design a quantization method for the prompts of vision-language models.  
    \item Following the principles we concluded, we build our method and successfully quantize the prompts as well as some other weights, and our method significantly outperforms existing ones as in \cref{fig:performance}. Extensive experiments show great power of our method. In base-to-new generalization, domain generalization, cross-dataset transfer and few-shot learning settings, we consistently reach competitive results, winning many state-of-the-art tuning methods with a much smaller size of model.
\end{itemize}

\section{Related Works}
\label{sec:related_works}
\subsection{Vision-Language Models}
In recent times, large-scale vision-language models have demonstrated remarkable performance across various tasks. Seminal works such as~\cite{radford2021learning,jia2021align,zhai2022lit,yao2022filip,yuan2021florence}. Classic works have focused on learning multimodal representations through self-supervised methods using extensive sets of image-text pairs. Among these, CLIP\cite{radford2021learning} stands out as a milestone achievement, employing contrastive learning to align vision and language representations and achieving exceptional performance. A well-trained vision-language model is invaluable, offering substantial support to various fields. Successful applications of these robust models include few-shot recognition~\cite{zhou2022conditional,zhou2021coop}, detection tasks~\cite{rasheed2022bridging,maaz2022class,feng2022promptdet,zang2022open}, and segmentation tasks~\cite{li2022languagedriven,rao2022denseclip,ding2022decoupling,luddecke2022image}. Furthermore, for video data, research efforts have emerged in video classification~\cite{qian2022multimodal} and video understanding~\cite{ju2022promptingvideounderstand}.

\subsection{Parameter-Efficient Fine-Tuning}
Recently, a series of works~\cite{hu2021lora,houlsby2019parameter,liu2020ptuning,li2021prefix,jia2022vpt,zhou2021coop,zhou2022conditional,lian2022ssf,hao2023consolidator,jie2023binarypet,hao2023repprompt,xiong2024pyra,lyu2024lftl} have been proposed to help transfer the learnt knowledge, where one of the most popular field is parameter-efficient fine-tuning (PEFT). PEFT aims to transfer a pre-trained model to downstream tasks by a minimum number of parameters. Originating from natural language processing tasks, classic methods like adapter~\cite{houlsby2019parameter}, prompt tuning~\cite{li2021prefix,lester2021power,liu2021p,shin2020autoprompt,liu2021pre,jiang2021can} and LoRA~\cite{hu2021lora} follows a similar principle to add extra modules with a small number of parameters into the backbone model, freeze the original parameters and only tune and store the newly added parameters. 
Inspired by the success of PEFT in language field, researchers have extended such kind of approaches to adapt visual models in a similar fashion~\cite{jia2022vpt,zhang2022noah,hao2023consolidator,lian2022ssf,chen2022adaptformer}. In the field of vision-language modeling, several explorations have been made as well. Bahng et al.\cite{bahng2022visual} exclusively apply prompt tuning to the image encoder. CoOp\cite{zhou2021coop} replaces the fixed template in CLIP~\cite{radford2021learning} with tunable text prompts. CoCoOp~\cite{zhou2022conditional} leverages image features to guide the optimization of tunable text prompts in CoOp. Other works~\cite{khattak2023maple,lee2023rpo} optimize both image and text prompts simultaneously and establish additional connections between different modalities. To mitigate overfitting and catastrophic forgetting, various works~\cite{khattak2023promptsrc,yao2023kgcoop,bulat2023lasp,zheng2023regularized} integrate regularization modules or losses into prompt tuning. 

\subsection{Quantization}
Quantization is one of the most effective compression methods~\cite{ding2021resrep,yu2022unified,bolya2023token,liu2017learning,li2016pruning,hao2023manipulating,ding2019centripetal} for deep learning models. Generally, parameters such as weights and activations are typically stored as 32-bit floating-point numbers, which consume a significant amount of memory and require intensive computation during inference. Quantization~\cite{han2015deep,gholami2022survey,nagel2021white} involves representing these parameters with reduced precision, such as 8-bit integers or even lower bit-widths. By doing so, quantization can significantly reduce the memory footprint and computational complexity of the model without significantly sacrificing accuracy. Quantization methods can be divided into two groups, Post-Training Quantization (PTQ)~\cite{yuan2022ptq4vit,liu2021post,hubara2021accurate,banner2019post,finkelstein2019fighting,meller2019same,nagel2019data,li2023repq} that consumes few resources but suffers higher accuracy loss, and Quantization-Aware Training (QAT)~\cite{nagel2022overcoming,bhalgat2020lsq+,esser2019learned,jung2019learning,li2022q} that relies on plenty of resources for training and shows better accuracy. Existing works aim to minimize quantization error to improve accuracy, while our work demonstrates that both excessive and insufficient errors are detrimental to model generalization. To achieve optimal generalization performance, a moderate error is required.

\section{Exploring Quantization in Model Generalization}
\label{sec:motivation}
\subsection{Preliminaries: Prompt Tuning of Vision-Language Models}
CLIP comprises a text encoder $\mathcal{L}$ and an image encoder $\mathcal{V}$. Typically, $\mathcal{L}$ is implemented as a language transformer, whereas $\mathcal{V}$ may be realized using either a convolutional neural network or a vision transformer. In this study, following the methodologies outlined by~\cite{zhou2021coop,zhou2022conditional}, we employ a ViT-B/16 model~\cite{dosovitskiy2020image} as the image encoder $\mathcal{V}$, except where otherwise specified. The subsequent sections will provide a brief overview of the methods used to prompt CLIP for prediction tasks.
\\
\textbf{Text Encoder} Consider a text encoder composed of $M$ layers. For the $k$-th layer, denoted as $\mathcal{L}_{k}$, the inputs consist of a sequence of prompt tokens $P^{l}_{k-1}$ and a \textit{[CLS]} token $c^{l}_{k-1}$, while the outputs are represented by $P^{l}_{k}$ and $c^{l}_{k}$. The initial inputs, $P^{l}_{0}$ and $c^{l}_{0}$, correspond to the word embeddings of the prompts combined with the label, such as ``\textit{A photo of a [CLS]}'' or alternatively, some randomly initialized vectors. Formally, we denote $P^{l}_{k} \in \mathbb{R}^{n^{l} \times d^{l}}$ and $c^{l}_{k} \in \mathbb{R}^{d^{l}}$, where $n^{l}$ signifies the length of the text prompts and $d^{l}$ represents the dimension of the word embeddings. For each layer, $1 \leq k \leq M$, the relationship is given by $[P^{l}_{k}, c^{l}_{k}] = \mathcal{L}_{k}([P^{l}_{k-1}, c^{l}_{k-1}])$.
The output feature of the text encoder $f^l\in \mathbb{R}^{d^v}$, where $d^v$ is the dimension of the visual feature space, is calculted by projecting the \textit{[CLS]} token of its last layer to the visual latent space through a linear transformation, \ie $f^{l}={\rm Proj}(c^{l}_{M})$.
\\
\textbf{Image Encoder} Suppose there are $N$ layers in the image encoder. For $k$-th layer $\mathcal{V}_{k}$, the inputs are a series of image tokens $I_{k-1}$, a classification token $c^{v}_{k-1}$ and prompt tokens $P^{v}_{k-1}$, and the outputs are $I_{k}$, $c^{v}_{k}$ and $P^{v}_{k}$. The inputs of the first layer $I_{0}$ and $c^{v}_{0}$ are the patch embeddings of the input image and the pre-trained class token. $P^{v}_{0}$ is randomly initialized. Formally, $I_{k}\in \mathbb{R}^{p\times d^v}$, $c^{v}_{k}\in \mathbb{R}^{d^v}$ and $P^{v}_{k}\in \mathbb{R}^{n^{v}\times d^{v}}$, where $p$ denotes the number of image patches and $d^{v}$ denotes the dimension of visual embedding. $\forall 1\leq k \leq N$, $[P^{v}_{k},c^{v}_{k},I_{k}]=\mathcal{V}_{k}([P^{v}_{k-1},c^{v}_{k-1},I_{k-1}])$. The output feature of the image encoder is $f^{v}=c^{v}_{N}$.
\\
\textbf{Prediction} For image classification, suppose there are $C$ classes, and $\{f^l_{c}\}_{c=1}^{C}$ are the text features. Label $y$'s probability is $p(y|f^v)=\frac{\mbox{exp}(\mbox{sim}(f^v,f^{l}_{y})/\tau)}{\sum_{c=1}^{C}\mbox{exp}(\mbox{sim}(f^v,f^{l}_{c})/\tau)}$ where sim$(\cdot,\cdot)$ denotes cosine similarity function and $\tau$ is temperature. The final prediction is $\hat{z}=\mathop{\arg\max}\limits_{1\leq y\leq C}(p(y|f^v))$.


We have introduced shallow prompts. There are also different types of prompts. Several works~\cite{jia2022vpt,khattak2023maple} use them for improve performance.They directly add and tune the prompt in each layer in the feature encoder, instead of inheriting the output prompt calculated by the last encoder. Now we have $[\_,c^{l}_{k}]=\mathcal{L}_{k}([P^{l}_{k-1},c^{l}_{k-1}])$ and $[\_,c^{v}_{k},I_{k}]=\mathcal{V}_{k}([P^{v}_{k-1},c^{v}_{k-1},I_{k-1}])$. Note that $P^l$ and $P^v$ are independent tunable parameters. They are no longer determined by the previous layer.


\subsection{Rethinking the Relationship between Noise and Generalization}
\label{subsec:rethinking_noise}

\begin{figure}[tb]
  \centering
  \begin{subfigure}{0.45\linewidth}
    \includegraphics[width=\linewidth]{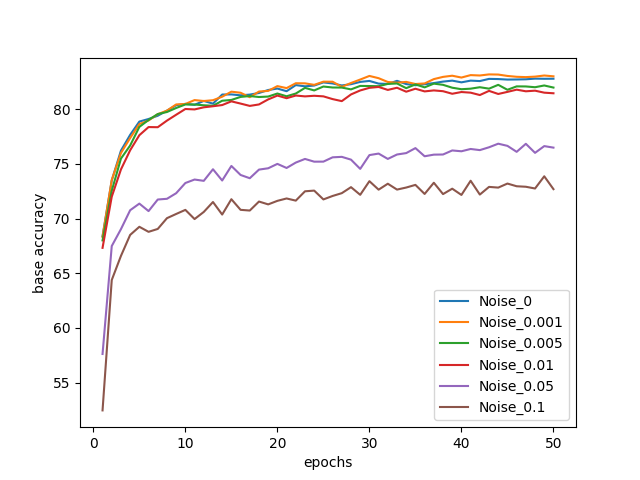}
    \label{fig:motivation_base_acc}
  \end{subfigure}
  \begin{subfigure}{0.45\linewidth}
    \includegraphics[width=\linewidth]{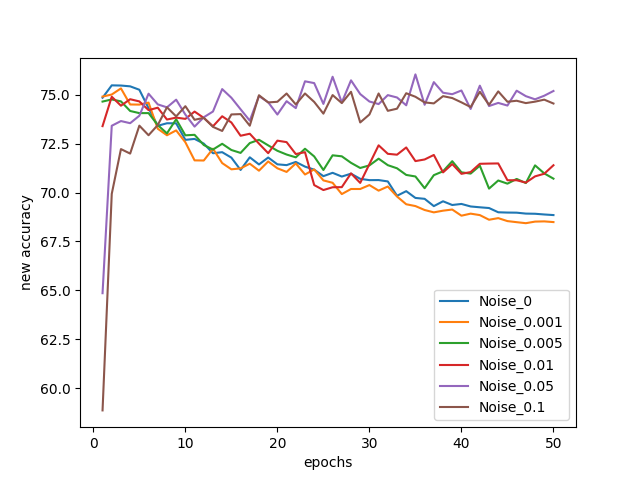}
    \label{fig:motivation_new_acc}
  \end{subfigure}
  \setlength{\abovecaptionskip}{-0.4cm}
  \setlength{\belowcaptionskip}{-0.3cm}
  \caption{The variations of the specialization capability represented by the test accuracy on base, seen classes and the generalization capability represented by the test accuracy on new, unseen classes on average of the ten datasets in base-to-new generalization setting in \cref{sec:experiments} except for time-consuming ImageNet. The curves of different colors represent the data under the influence of random Gaussian noise of different intensities, \eg ``Noise\_0.01'' adds random noise with a distribution of $\mathcal{N}(0, 0.01^{2})$ to the prompt. ``Noise\_0'' denotes the baseline prompt tuning. As training progresses, the generalization capability of baseline prompt tuning continuously decreases while the specialization capability improves. Therefore, we expect that adding noise can achieve a better balance between generalization and specialization. However, excessive noise, \eg 0.1, greatly diminishes the model's specialization capability, while insufficient noise, \eg 0.001, fails to provide effective regularization. Only noise of moderate intensity outperforms baseline in specialization-generalization trade-off, effectively enhancing the unseen class accuracy without significantly compromising seen class accuracy.}
  \label{fig:motivation}
\end{figure}

Some existing work has explored how to use the addition of noise to suppress model overfitting, \eg, techniques like Dropout~\cite{srivastava2014dropout} that randomly drops connections, and random jittor that directly introduce perturbations in the input data. However, the impact of directly adding noise to model weights has been less explored, especially in the context of prompt structures, which have become popular only in recent years, in Transformer architectures. 
The variations of the specialization capability represented by the test accuracy on base, seen classes and the generalization capability represented by the test accuracy on new, unseen classes. The curves of different colors represent the data under the influence of random Gaussian noise of different intensities, \eg ``Noise\_0.01'' adds random noise with a distribution of $\mathcal{N}(0, 0.01^{2})$ to the prompt. ``Noise\_0'' denotes the baseline prompt tuning. As training progresses, the generalization capability of baseline continuously decreases while the specialization capability improves. Therefore, we expect that adding noise can achieve a better balance between generalization and specialization. However, excessive noise, \eg 0.1, greatly diminishes the model's specialization capability, while insufficient noise, \eg 0.001, fails to provide effective regularization. Only noise of moderate intensity outperforms baseline in specialization-generalization trade-off, effectively enhancing the unseen class accuracy without significantly compromising seen class accuracy.

Quantization, the technique of mapping parameter values from high precision to low precision, can also be viewed as introducing some form of noise into the parameters, and thus can possibly improve the genralizability as the noise did in \cref{fig:motivation}. Compared to Gaussian noise, quantization error is more controllable. Better still, quantization also has another major advantage: it can significantly reduce the storage required for parameters. Therefore, using quantization algorithms to generalize vision-language models is a very promising direction.


\begin{figure}[tb]
	\centering
	\begin{minipage}[t]{0.192\linewidth}
	\centering
	\includegraphics[width=\linewidth]{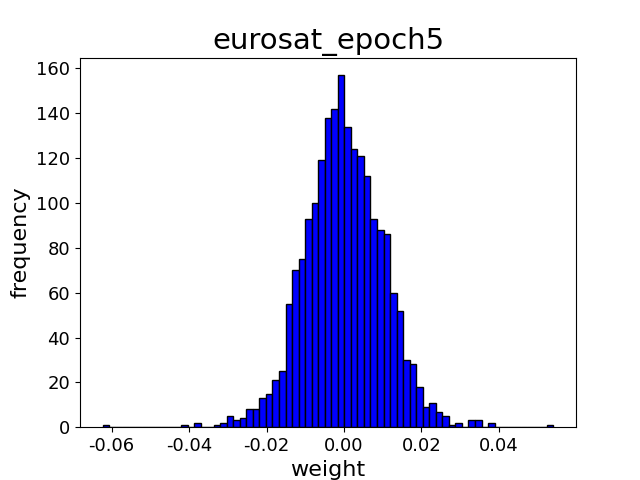} \\
        \includegraphics[width=\linewidth]{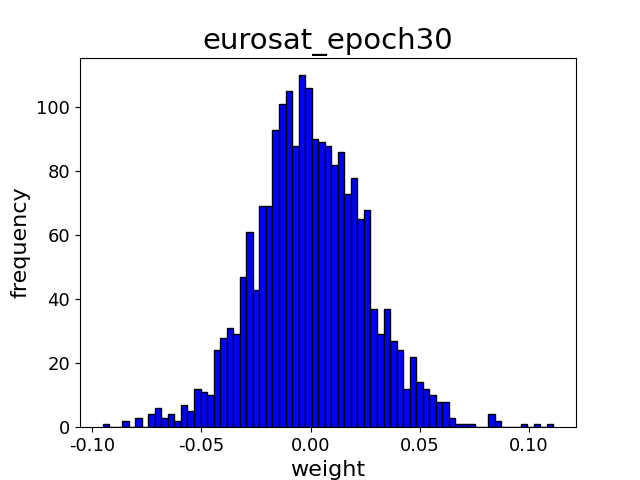}
	\end{minipage}
	\begin{minipage}[t]{0.192\linewidth}
	\centering
	\includegraphics[width=\linewidth]{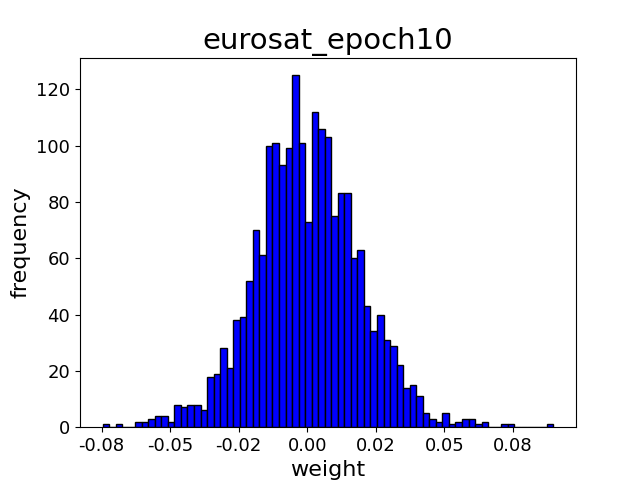} \\
        \includegraphics[width=\linewidth]{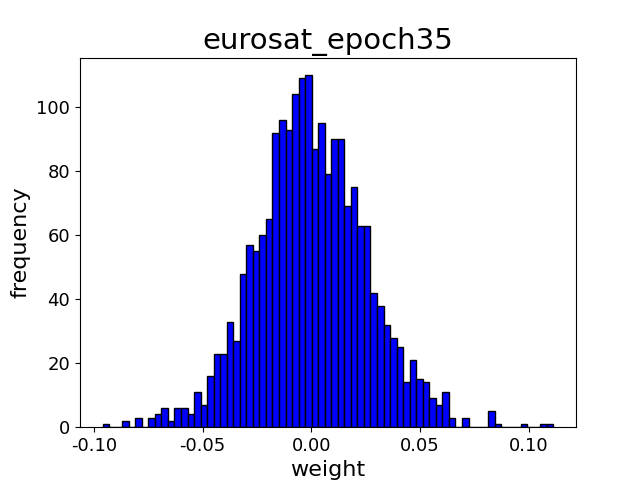}
	\end{minipage}
	\begin{minipage}[t]{0.192\linewidth}
	\centering
	\includegraphics[width=\linewidth]{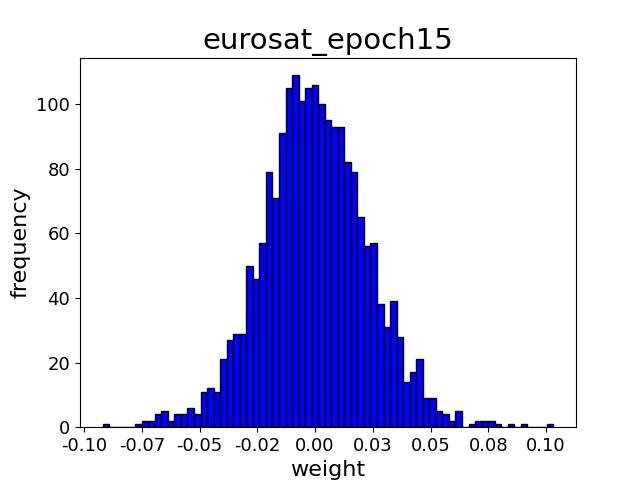} \\
        \includegraphics[width=\linewidth]{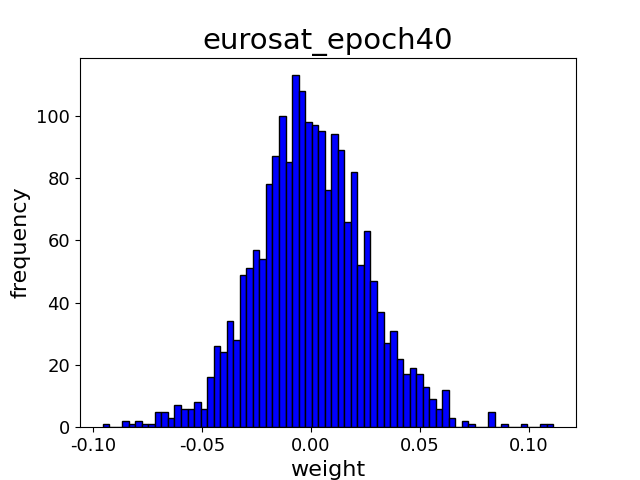}
	\end{minipage}
 	\begin{minipage}[t]{0.192\linewidth}
	\centering
	\includegraphics[width=\linewidth]{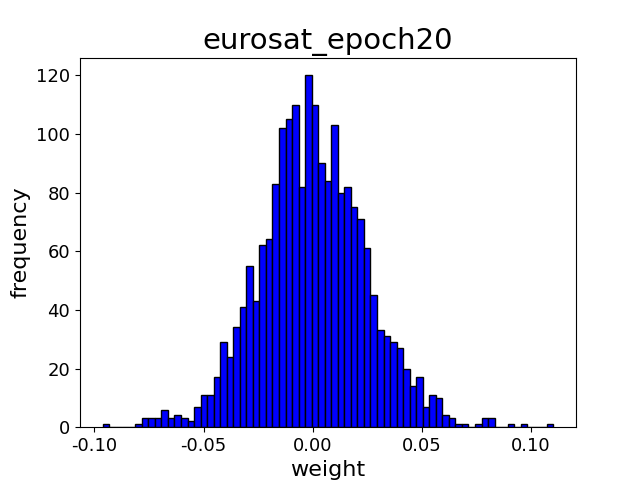} \\
        \includegraphics[width=\linewidth]{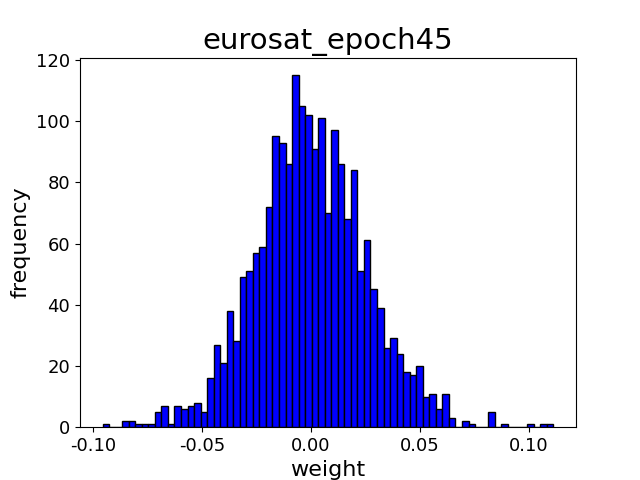}
	\end{minipage}
 	\begin{minipage}[t]{0.192\linewidth}
	\centering
	\includegraphics[width=\linewidth]{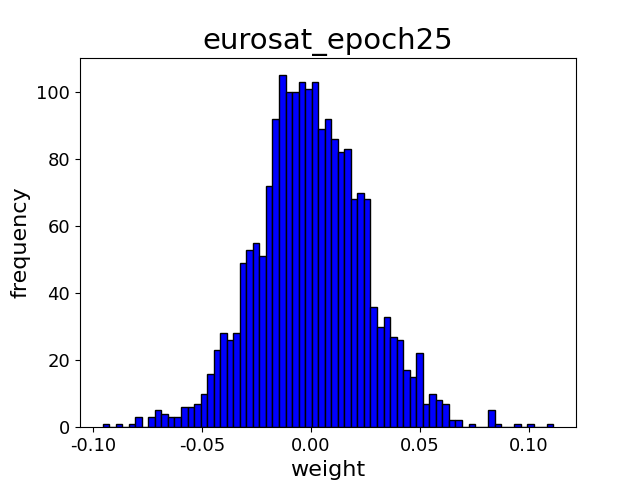} \\
        \includegraphics[width=\linewidth]{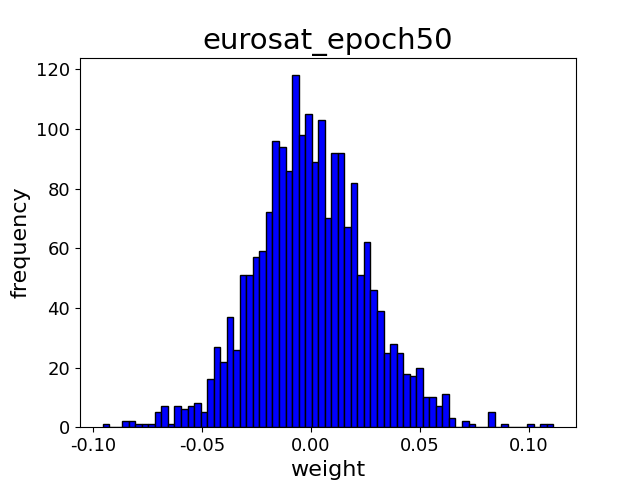}
	\end{minipage}
 
 \setlength{\abovecaptionskip}{-0.01cm}
  \setlength{\belowcaptionskip}{-0.3cm}
  \caption{The histogram about the frequency of weights of prompts during training of CoOp~\cite{zhou2021coop} on eurosat dataset. We find the shape of the prompt's weight distribution remains largely unchanged throughout the entire training, but the variance of the weight distribution increases rapidly at the beginning of training. Additionally, we also notice that there are almost no outliers in the prompt's weights throughout the entire training process, and apart from the unstable initial training phase, the changes in weights between adjacent phases are not significant, indicating a very gentle overall updating trend. }
  \label{fig:hist}
\end{figure}

\subsection{Characteristics of Prompts in Vision-Language Models}
\label{subsec:characteristics}
In this subsection, we will demonstrate the characteristics of prompts in the vision-language model to facilitate targeted design of quantization schemes. 

We start from analyzing the training procedure of CoOp~\cite{zhou2021coop}, which freezes all the parameters in backbone and tunes the added prompts only. The histogram about the frequency of weights of prompts for different training epochs is shown in \cref{fig:hist}. We observe that the shape of the prompt's weight distribution remains mostly consistent throughout the entire training process. However, the variance of the weight distribution increases rapidly at the initial stages of training. Refer to \cref{fig:KLD_eurosat} for more details. Moreover, we notice minimal presence of outliers in the prompt's weights throughout the training process. Apart from the unstable initial training phase, the changes in weights between consecutive phases are not substantial, indicating a very gentle overall updating trend.

Taking into account the observations from \cref{subsec:characteristics,subsec:rethinking_noise}, we can draw the following conclusions: 
\begin{itemize}
    \item Because the prompt weights are not sensitive to noise, and moderate noise in the current scenario not only does not degrade the model's performance but actually enhances its generalization ability, we can possibly adopt some quantization strategies that are considered very aggressive and highly likely to significantly degrade performance, such as 1-bit quantization, without causing destructive effects on the model's generalization ability.
    \item Given that the parameters targeted for quantization in the current scenario constitute only a tiny fraction of the total model parameters, we speculate that QAT may have more advantages over PTQ, as any additional training costs incurred by QAT which is considered a major drawback in traditional situations would naturally be kept at a low level due to the low proportion of parameters targeted for quantization. Due to the observation in \cref{fig:motivation} that both excessive and insufficient noise are detrimental to generalization, in designing the QAT algorithm, we no longer need to always minimize quantization error as the objective, as in classic algorithms. Instead, we aim to maintain the quantization error at a moderate level. Furthermore, as seen in \cref{fig:hist}, the changes in prompt weight distribution during most of the training time are gentle, indicating that we can appropriately continue to use past-time algorithm state without the need for real-time updates.
    \item Throughout the training process, the shape of the prompt weight distribution remains roughly unchanged, with the main variation coming from the distribution's variance. Therefore, it may be beneficial to attempt to eliminate the translation and scaling transformations of the distribution by normalizing the distribution before quantization and denormalizing it after quantization, which might help improve quantization accuracy. 
    \item Since there are almost no outliers throughout the entire training process, some classic algorithms are revitalized. Specifically, clustering algorithms like K-Means are good choices, because one of the major drawbacks of them in traditional usage is that they are greatly affected by outliers. In addition, as we have analyzed before, there is not a high demand for dynamically minimizing quantization error in vision-language generalization, opting for a more efficient non-parametric clustering approach seems a better choice compared with some complex parametric approaches.
\end{itemize}

\section{Prompt Quantization}
\label{sec:promptquant}
\subsection{Preliminaries: Quantization Basis}
Suppose there are $m$ parameters in total to be quantized, which are denoted by $W\in \mathbb{R}^{m}$. Each $w_{i}$ here is a high-precision float-type number. A $b$-bit model quantization algorithm divides the value range of parameters into $2^b$ intervals $\{\mathcal{U}_{i}, 1\leq i \leq 2^b\}$, aiming to find a mapping $Q$ from intervals to points, mapping all values within an interval $\mathcal{U}_{i}$ to the same quantized value $q_{i}$, \ie $Q(x)=q_{i},\forall x\in \mathcal{U}_{i}$. We define the quantization error $E=\sum_{i=1}^{N}{||Q(W_{i})-W_{i}||^{2}}$. Such error serves as the objective of K-Means algorithm as well,  so it can be ensured that each time clustering is redone using the K-Means algorithm, the quantization error will most likely decrease.

\begin{figure}[tb]
  \centering
    \includegraphics[width=0.8\linewidth]{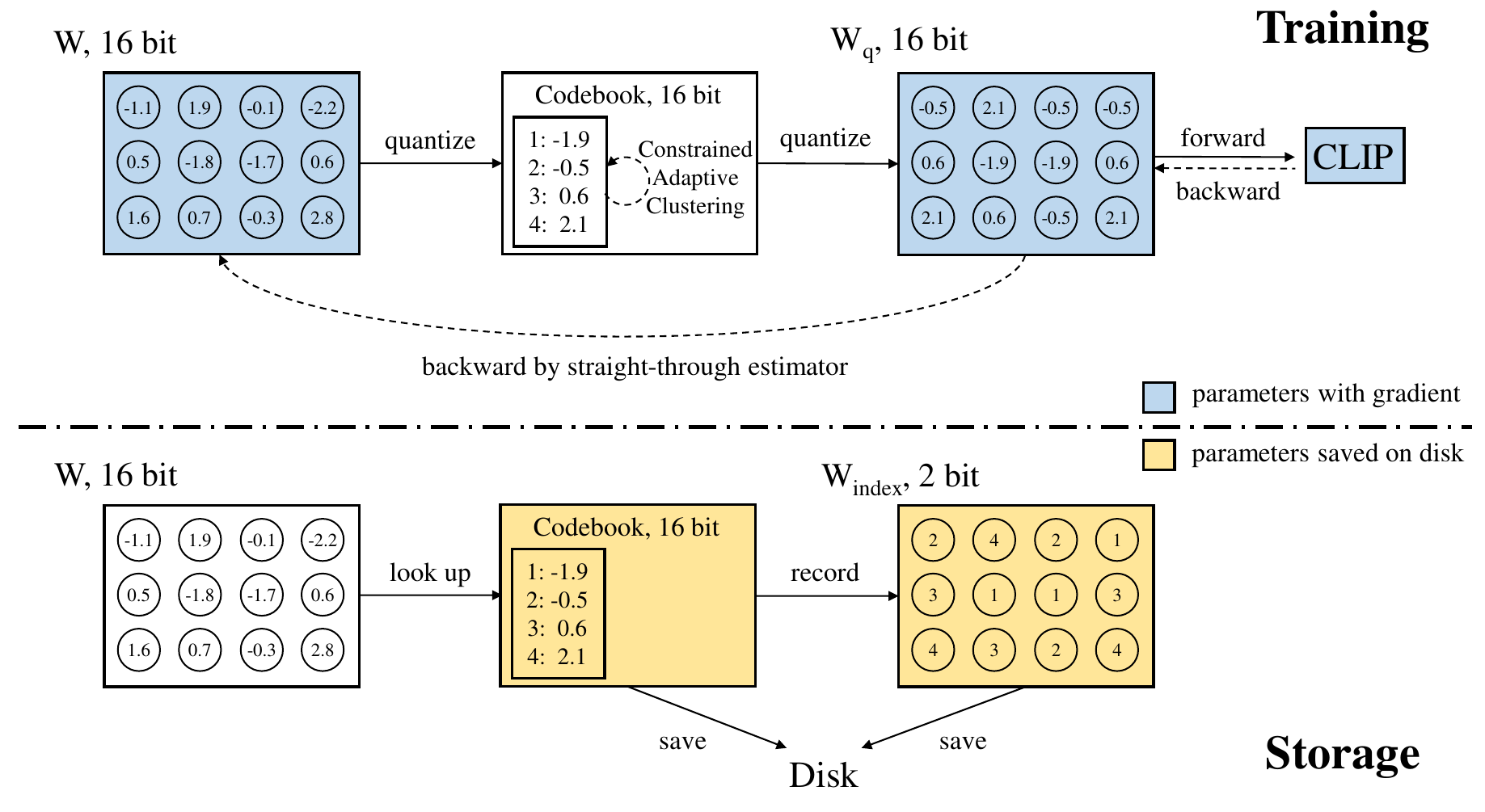}
    \label{fig:motivation_base_acc}
  \setlength{\abovecaptionskip}{-0.1cm}
  \setlength{\belowcaptionskip}{-0.3cm}
  \caption{Overview of Quantized Prompt. We set $b=2$ for a clear explanation. Normalization and denormalization processes are not shown here.}
  \label{fig:QPrompt}
\end{figure}

\subsection{Overview of Quantized Prompt}
Based on the observation and analysis in \cref{subsec:characteristics}, we build a simple yet quite proper quantization algorithm for generalizing vision-language models. As stated before, we choose a widely-used clustering method K-Means, to construct the mapping $Q$. Initially, K-Means algorithm is run to fit the pre-trained prompt weights, and record the codebook, \ie cluster centers. Normalization and denormalization are conducted beyond quantization to alleviate the influence of varying variance. Formally, given the original prompt $W$ with $N$ parameters, we first calculate $\mu=\frac{1}{N}\sum_{i=1}^{N}{W_{i}}$ and $\sigma=\sqrt{\frac{\sum_{1\leq i \leq N}{(W_{i}-\mu)^{2}}}{N}}$, then normalize $W$ and get $\hat{W}=\frac{W-\mu}{\sigma}$. K-Means quantization is applied on $\hat{W}$, and the quantized prompt $W_{q}=\sigma Q(\hat{W})+\mu$. Since $Q$ is not differentiable, we adopt a common practice to directly propagate the gradient across the quantization function by Straight-Through Estimator (STE), \ie $\frac{\partial Q(x)}{\partial x}=x$. 

The overall framework is shown in \cref{fig:QPrompt}. In training, we keep tuning the weight by the fake gradient propagated from quantization operation. For storage, we first convert the fp16 parameters in prompt to $b$-bit indexes, which could be used to search for the corresponding cluster center in the codebook. In \cref{fig:QPrompt}, we set $b=2$ for a clear explanation. Compared with baseline method, our quantized method could save a lot more storage space. Specifically, for storage, an ordinary method needs $16N$ bits, while ours only needs $bN+2^b\times16$. In experiments, we usually set $b$ to $1$, $2$ or $4$, which is far more smaller than $N$. For example, with $b=1$, the storage space of our method is roughly \textbf{$16\times$} smaller than baseline.

Note that the codebook here is updated by a rule called ``constrained adaptive clustering''. We will give a detailed description of it in the following subsection.

\subsection{Constrained Adaptive Clustering for Quantization}
\label{subsec:updaterule}
In \cref{fig:QPrompt}, re-clustering the prompt every iteration like classic QAT algorithms did to keep the codebook updated at all times is not a good choice. There are three reasons: 1. K-Means algorithm is not quite efficient, and thus if we run it every iteration then the training efficiency would decrease; 2. Only moderate noise promotes generalization, so keep updating the clusters to minimize the quantization error may be not helpful but harmful; 3. From \cref{fig:hist}, we can observe that the weight changes are very gentle for most of the training time, and the weight distributions of adjacent stages are highly similar. Therefore, even if we update the internal clustering state of K-Means with these already very similar data at each iteration, it's difficult to generate a better clustering solution, which is just futile effort.

We propose constrained adaptive clustering to instruct the update of parameters inside K-Means which would not updated by gradient at all. Intuitively, first, we do not want too often update, so we set a minimum cluster update interval $t$. Second, to avoid K-Means meaninglessly handling similar weights with current cached state, \ie the weights of prompt that triggers the last re-clustering, we plan to only do re-clustering when current weight distribution is far different from the cached weight distribution. To reach the goal, Kullback-Leibler divergence is a good metric. However, if we want to compute the Kullback-Leibler divergence between two sets of discrete random variables, they must be defined on the same event space. However, that's not the case. So we 
first project each of them into the same event space spanning by the K-Means clustering algorithm. Specifically, given a current weight $W^{cur}$, cached weight $W^{old}$, quantization function $Q$, cluster centers $\mathcal{C}=\{c_{i},1\leq i\leq 2^{b}\}$, we will compute the $W_{index}^{cur}$ and $W_{index}^{old}$ as in \cref{fig:QPrompt}. 
Then, we can obtain the probability distributions of the indices $p^{cur}$ and $p^{old}$, respectively, implied by $W_{index}^{cur}$ and $W_{index}^{old}$, thus successfully transforming two originally unrelated discrete random variables into the same event space. The Kullback-Leibler divergence could be computed as follows:

\begin{equation}
    KL(p^{cur}||p^{old})=\sum_{i=1}^{2^{b}}{p^{cur}(i)\log{\frac{p^{cur}(i)}{p^{old}(i)}}}
    \label{eq:KL}
\end{equation}

The update would continue only if Kullback-Leibler divergence exceed a certain threshold $T_{KL}$ which shows the distribution difference is significant.

\section{Experiments}
\label{sec:experiments}


We evaluate the method and make comparisons with the latest state-of-the-art methods in terms of the following settings across a wide range: 
\begin{enumerate}
    \item \textbf{Base-to-new generalization}, which models are trained with base classes and evaluated on both base and new classes.
    \item \textbf{Domain generalization}, which aims to show generalization to the domain shift, especially for out-of-distribution data. 
    \item \textbf{Cross-dataset transfer}, which aims to see if the method has the potential to transfer beyond a single dataset. It is a much more challenging problem because the fundamentals can be totally changed across different datasets.
    \item \textbf{Few-shot learning}, which aims to evaluate the adaptation performance of the model to extract knowledge from a dataset whose samples are very few, \eg 1, 2, 4, 8 or 16 samples.
\end{enumerate}

All methods are initialized with the same CLIP weights provided by the open-source CLIP~\cite{radford2021learning}. In Appendix, we provide more details of datasets, experimental setting and competitors introduction. Due to page limit, we also put some detailed experimental results into Appendix, \eg the performance of each method on each dataset in base-to-new generalization setting.

\begin{table}[tb]
    \scriptsize
    \setlength{\belowcaptionskip}{-0.05cm}
    \setlength{\tabcolsep}{1.5mm}
    \caption{Comparisons with latest methods in base-to-new generalization. H: harmonic mean~\cite{xian2017zero}. QCoOp and QMaPLe are significantly better than other competitors, which prove that our quantization design could be fruitfully integrated into many existing approaches to further improve the performance and efficiency.}
    \label{tab:results_base_to_new_generalization}
    
    \centering
    \begin{tabular}{lccc|c}
    \toprule
    & Size & Base & New & H \\
    \midrule
    CLIP~\cite{radford2021learning} & 0KB & 69.34 & 74.22 & 71.70 \\
    CoOp~\cite{zhou2021coop} & 4.1KB & 82.69 & 63.22 & 71.66 \\
    CoCoOp~\cite{zhou2022conditional} & 70.8KB & 80.47 & 71.69 & 75.83 \\
    Adapter~\cite{gao2021clip-adapter} & 1051KB & 82.62 & 70.97 & 76.35 \\
    LoRA~\cite{hu2021lora} & 258KB & \textbf{84.30} & 67.33 & 74.86 \\
    ProGrad~\cite{zhu2023prograd} & 16.4KB & 82.79 & 68.55 & 75.00 \\
    \rowcolor{tabhighlight}
    QCoOp & \textbf{0.26KB} & 80.68 & \textbf{74.44} & \textbf{77.43} \\
    \hline
    MaPLe~\cite{khattak2023maple} & 7096KB & 82.28 & 75.14 & 78.55 \\
    \rowcolor{tabhighlight}
    QMaPLe & \textbf{1774KB} & \textbf{83.02} & \textbf{75.57} & \textbf{79.12} \\
    \bottomrule
    \end{tabular}
\end{table}

\subsection{Main Results}
\label{subsec:main_result}
\subsubsection{Base-to-new Generalization}

The average results over 11 datasets are shown in \cref{tab:results_base_to_new_generalization}. For complete results, please refer to the Appendix. Our proposed QCoOp reaches \textbf{77.43\%} average harmonic mean accuracy within merely \textbf{0.26KB} size. QCoOp outperforms all kinds of lightweight state-of-the-art methods with much more efficiency and higher accuracy. For heavier methods like MaPLe, our method can be fruitfully integrated into existing solutions. Besides prompts, we also perform a similar quantization operation on the other weights of MaPLe that are in the linear layers. As a result, QMaPLe not only shows stronger generalization and adaptation capability and gives 0.57\% higher accuracy, but also enjoys a much more smaller model size compared with the original MaPLe. 

Notably, when comparing QCoOp and a lightweight method ProGrad, even if QCoOp is $63\times$ smaller than ProGrad, QCoOp still outperforms it by a clear margin, demonstrating the outstanding efficiency and effectiveness.

\begin{table}[tb]
\scriptsize
\setlength{\tabcolsep}{0.8mm}
\centering
\setlength{\belowcaptionskip}{-0.05cm}
\caption{Comparisons with latest methods in domain generalization. QCoOp gets comparable or even better results with the latest state-of-the art methods with much fewer parameters, showing excellent robustness for domain shift. }
\label{tab:domain_generalization}
\begin{tabular}{lc|c|p{1.6cm}<{\centering}p{1.6cm}<{\centering}p{1.8cm}<{\centering}p{1.6cm}<{\centering}|c}
\toprule
& & \multicolumn{1}{c}{\textbf{Source}} \vline & \multicolumn{4}{c}{\textbf{Target}} \vline \\
& Size & ImageNet & -V2 & -Sketch & -Adversarial & -Rendition & Average \\
\midrule
CLIP  & 0KB & 66.73 & 60.83 & 46.15 & 47.77 & 73.96 & 57.18 \\
CoOp  & 4.1KB & 71.51 & 64.20 & 47.99 & 49.71 & 75.21 & 59.28 \\
CoCoOp  & 70.8KB & 71.02 & 64.07 & 48.75 & 50.63 & 76.18 & 59.91 \\
Adapter & 1051KB & 69.33 & 62.53 & 47.67 & 49.17 & 75.42 & 58.70 \\
LoRA & 258KB & 70.30 & 62.37 & 42.43 & 38.40 & 68.97 & 53.04 \\
ProGrad  & 16.4KB & \textbf{72.24} & \textbf{64.73} & 47.61 & 49.39 & 74.58 & 59.07 \\
\rowcolor{tabhighlight}
QCoOp & 0.26KB & 70.67 & 63.87 & \textbf{48.93} & \textbf{51.10} & \textbf{76.90} & \textbf{60.20} \\
\bottomrule
\end{tabular}
\end{table}

\subsubsection{Domain Generalization}
In this paragraph, ImageNet, ImageNet-A, ImageNet-R, ImageNet-v2, and ImageNet-S are used to construct domain generalization experiments. As shown in \cref{tab:domain_generalization}, on downstream target datasets, QCoOp gets better average accuracy compared with the other methods with significantly better efficiency. For CLIP, CoOp, CLIP-Adapter and ProGrad, there is a clear performance gap between our method and them.

\begin{table}[tb]
\scriptsize
\centering
\setlength{\belowcaptionskip}{-0.05cm}
\caption{Results in the cross-dataset transfer setting. QCoOp gives the highest accuracy on 5 of 10 datasets, which well demonstrates that QCoOp could maximally extract general and data-agnostic knowledge from input images.}
\label{tab:crossdataset}
\begin{tabular}{lc|c|cccccccccc|c}
\toprule
& & Source & \multicolumn{10}{c}{Target} \vline &\\
& \rotatebox{90}{Size} & \rotatebox{90}{ImageNet} & \rotatebox{90}{Caltech101} & \rotatebox{90}{Pets} & \rotatebox{90}{Cars} & \rotatebox{90}{Flowers} & \rotatebox{90}{Food101} & \rotatebox{90}{Aircraft} & \rotatebox{90}{Sun397} & \rotatebox{90}{DTD} & \rotatebox{90}{EuroSAT} & \rotatebox{90}{UCF101} & \rotatebox{90}{Average}\\
\midrule
CoOp & 4.1KB & \textbf{71.51} & 93.70 & 89.14 & 64.51 & 68.71 & 85.30 & 18.47 & 64.15 & 41.92 & 46.39 & 66.55 & 63.88\\
CoCoOp & 70.8KB & 71.02 & \textbf{94.43} & 90.14 & 65.32 & \textbf{71.88} & 86.06 & 22.94 & \textbf{67.36} & \textbf{45.73} & 45.37 & 68.21 & 65.74\\
Adapter & 1051KB & 69.33 & 93.43 & 88.87 & 64.40 & 70.27 & 85.63 & \textbf{24.67} & 65.80 & 44.90 & 47.70 & 66.00 & 65.17\\
\rowcolor{tabhighlight}
QCoOp & 0.26KB & 70.63 &  94.07 &  \textbf{90.53} & \textbf{65.97} & 71.33 & \textbf{86.23} & 22.73 & 66.80 & 44.20 & \textbf{48.23} &\textbf{69.17}  & \textbf{65.93}\\
\bottomrule
\end{tabular}
\end{table}

\subsubsection{Cross-dataset Transfer}
In this paragraph, we do cross-dataset transfer evaluation to further verify our QPrompt. Results are shown in \cref{tab:crossdataset}. CoOp is good on source domain but fails on target domains. Probably because it focus too much on the dataset shown to its eyes and face overfitting and catastrophic forgetting problems, which leads to a severe performance drop on unseen objects. QCoOp wins on 5 of 10 datasets and its average accuracy is also slightly better than the best competitor CoCoOp, showing that QCoOp could maximally extract both general and data-agnostic knowledge from given images.


\begin{figure}[tb]
\centering
\begin{minipage}[t]{0.45\textwidth}
\centering
    \includegraphics[width=0.95\linewidth]{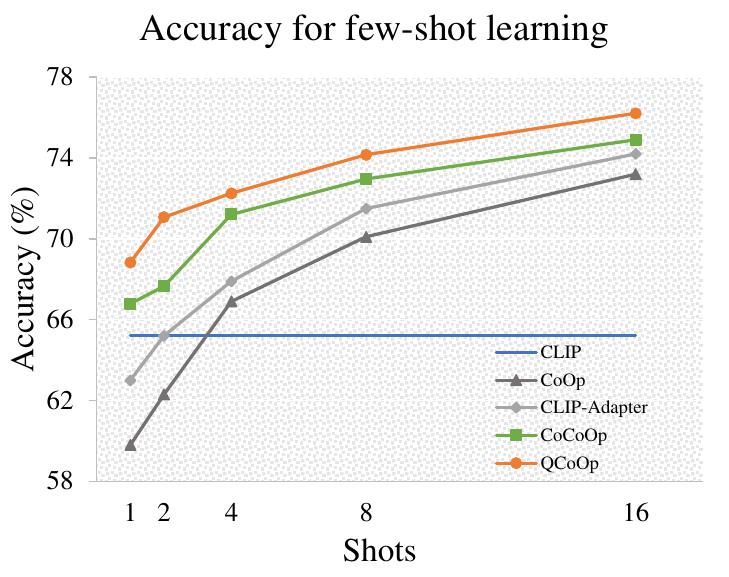}
  \setlength{\abovecaptionskip}{-0.05cm}
  \setlength{\belowcaptionskip}{-0.3cm}
  \caption{Average few-shot learning results on 11 datasets.}
  \label{fig:fewshot}
\end{minipage}
\begin{minipage}[t]{0.5\textwidth}
  \centering
    \includegraphics[width=0.95\linewidth]{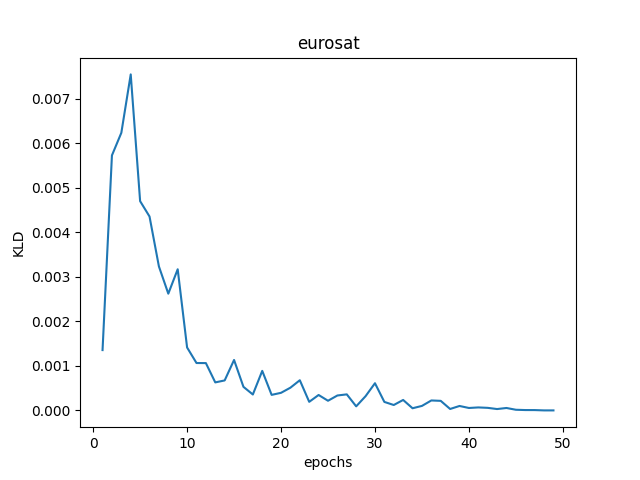}
  \setlength{\abovecaptionskip}{-0.02cm}
  \setlength{\belowcaptionskip}{-0.02cm}
  \caption{KLD trend of prompt weights under the same experiment with \cref{fig:hist}.}
  \label{fig:KLD_eurosat}
\end{minipage}
\end{figure}

\subsubsection{Few-shot Learning}
Here we will show the experiment results of QCoOp in the few-shot learning setting that is originated from CoOp. Seen from \cref{fig:fewshot}, QCoOp consistently outperforms zero-shot CLIP, CoOp, and CLIP-Adapter across all the shot numbers. Such results demonstrate the superiority of QCoOp in adaptation ability when there are few samples in downstream tasks.

Overall, in base-to-new generalization, domain generalization, cross-dataset transfer and few-shot learning, the proposed method can consistently accomplish state-of-the-art performance while enjoying extremely high parameter efficiency, fruitfully demonstrating the effectiveness and efficiency of the proposed method.

\section{Analysis}
\label{sec:analysis}

\subsection{Ablation Studies}
\label{subsec:ablation}

\begin{table}[tb]
    \centering
    \scriptsize
    \setlength{\belowcaptionskip}{-0.05cm}
    \caption{Ablation study on separate component of QCoOp. Norm: Normalization/Denormalization. CAC: Constrained Adaptive Clustering.}
    \setlength{\tabcolsep}{1.3mm}{
    \renewcommand{\arraystretch}{1.2}
    \begin{tabular}{|c|p{1.1cm}<{\centering}p{1.1cm}<{\centering}p{1.1cm}<{\centering}|ccc|}
        \hline
        & K-Means & Norm & CAC & base & new & H \\
        \hline
        \multirow{3}{*}{QCoOp}& \ding{51} & \ding{55} & \ding{55} & 78.71 & 72.55 & 75.50\\
        & \ding{51} & \ding{51} & \ding{55} & \textbf{78.84} & 73.09 & 75.85\\
        & \ding{51} & \ding{51} &\ding{51} & 78.24 &\textbf{74.02}& \textbf{76.07} \\
        \hline
    \end{tabular}
    }
    \label{tab:module ablation}
\end{table}

\paragraph{Component} In this subsection, we decompose the method into pieces and show the influence of each component. As in \cref{tab:module ablation}, we clearly show how much the K-Means algorithm, normalization/denormalization and constrained adaptive clustering influence the final performance. One interesting point is that K-Means+Norm+CAC only improves the new accuracy compared with K-Means+Norm, showing the superiority of our constrained adaptive clustering. 

\paragraph{Trend of the distributions of prompt between adjacent epochs.} To verify the opinions we proposed at the end of \cref{sec:analysis}, we show the KLD of prompt distributions between adjacent epochs during CoOp's training in \cref{fig:KLD_eurosat}. Clearly, this trend is consistent with what we summarized before.

\begin{table}[tb]
    \centering
    \scriptsize
    \setlength{\belowcaptionskip}{-0.05cm}
    \caption{Comparisons between QAT and PTQ based on CoOp. Both QAT and CoOp are trained with the same hyper-parameters. Here the quantization bit is 1.}
    \setlength{\tabcolsep}{2.7mm}{
    \renewcommand{\arraystretch}{1.2}
    \begin{tabular}{|c|c|c|c|c|}
        \hline
         & base & new & H \\
        \hline
        QAT & 80.72 & \textbf{72.35} & \textbf{76.31} \\
        \hline
        PTQ & \textbf{82.21} & 68.50 & 74.73\\
        \hline
    \end{tabular}
    }
    \label{tab:ptq_qat}
\end{table}

\paragraph{QAT v.s. PTQ} In \cref{tab:ptq_qat}, we study the choice of QAT or PTQ following the same strategy, K-Means clustering, in the paper. Results show that QAT consistently outperforms PTQ with the same hyper-parameters. The accuracy on unseen new classes of PTQ is significantly lower than QAT, again proving our opinion that quantization helps generalization by alleviating overfitting as well as catastrophic forgetting. Quantization error is not always undesirable.

\begin{table}[tb]
    \centering
    \scriptsize
    \setlength{\belowcaptionskip}{-0.05cm}
    \caption{Results across different quantization bits.}
    \setlength{\tabcolsep}{2.7mm}{
    \renewcommand{\arraystretch}{1.2}
    \begin{tabular}{|c|c|c|c|c|}
        \hline
         & Size & base & new & H \\
        \hline
        QCoOp ($b=1$)& 0.26KB & 79.49 &	72.65 &	75.92\\
        \hline
        QCoOp ($b=2$)& 0.52KB & 80.30 & 71.98	& 75.91\\
        \hline
        QCoOp ($b=4$)& 1.05KB & 80.92 & 71.22 &	75.76\\
        \hline
    \end{tabular}
    }
    \label{tab:ablation_bit}
\end{table}

\begin{figure}[tb]
  \centering
  \begin{subfigure}{0.45\linewidth}
    \includegraphics[width=\linewidth]{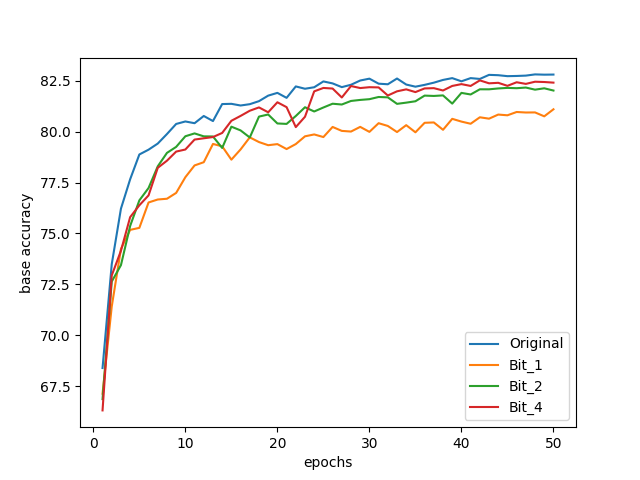}
    \label{fig:qmotivation_base_acc}
  \end{subfigure}
  \begin{subfigure}{0.45\linewidth}
    \includegraphics[width=\linewidth]{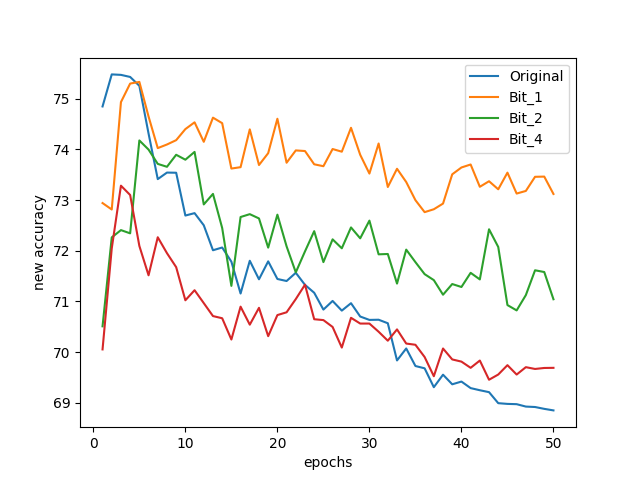}
    \label{fig:qmotivation_new_acc}
  \end{subfigure}
  \setlength{\abovecaptionskip}{-0.4cm}
  \setlength{\belowcaptionskip}{-0.3cm}
  \caption{The same training curves with \cref{fig:motivation}. Clearly, increasing quantization noise leads to the same phenomenon with increasing gaussian noise, \ie more generalizability (new accuracy) and less adaptability (base accuracy).}
  \label{fig:qmotivation}
\end{figure}

\paragraph{Quantization bit}
In \cref{tab:ablation_bit,fig:qmotivation}, we show the results across multiple quantization bits. In conclusions, more bits did not always lead to good results, and new accuracy continues decreasing as the training goes on.

\subsection{Performance on Self-Supervised Vision-Language Model}
In this paragraph, we explore the usage of QCoOp among other different backbones besides CLIP. We choose a self-supervised vision-language model, SLIP~\cite{mu2022slip} to further verify the universality and robustness of our proposed method. The experimental results are shown in \cref{tab:slip}. We could see that the base accuracies of QCoOp and CoOp are similar, but the new accuracies of QCoOp is much higher than CoOp. Such observation verifies our assumption that quantization is quite helpful for generalization again.

\begin{table}[tb]
    \centering
    \scriptsize
    \setlength{\belowcaptionskip}{-0.05cm}
    \caption{Results based on SLIP~\cite{mu2022slip} model. }
    \setlength{\tabcolsep}{2.7mm}{
    \renewcommand{\arraystretch}{1.2}
    \begin{tabular}{|c|c|c|c|c|}
        \hline
         & Size & base & new & H \\
        \hline
        CoOp & 4.1KB & \textbf{68.65} & 46.60 & 55.51 \\
        \hline
        QCoOp & 0.26KB & 68.33 & \textbf{74.04} & \textbf{71.07}\\
        \hline
    \end{tabular}
    }
    \label{tab:slip}
\end{table}


\section{Conclusion}
With the development of huge vision-language models, how to effectively and efficiently adapt such huge models to downstream tasks becomes a challenging problem. Much effort has been made to leverage the potential of prompt tuning in adapting vision-language models. However, existing methods suffer from inefficiency. To reach extremely efficient generalization, we propose QPrompt based on the detailed analysis and deep understanding of the characteristics of the prompt weight distribution. Following the several principles concluded by us, we use K-Means clustering algorithm as the base of our quantization method. To adaptively control the quantization error and minimize the number of re-clustering operations, we propose to do a dynamic check every few iterations. If the Kullback-Leible divergence between the current weights and the original weights used in the last clustering exceeds a predefined threshold and the quantization error also increases, we will let the model re-cluster the weights and update the new centers and weights. Similarly, continue this process until completion. The proposed method could be simply integrated into many of the existing vision-language tuning methods like CoOp and MaPLe and reach good performance. Importantly, our proposed method is significantly effective and efficient. Extensive experiments show that the designed quantization algorithm indeed improves the genralizability of vision-language model, and save a lot more storage space as well.

\paragraph{Acknowledgements.} This work was supported by National Natural Science Foundation of China (Nos. 61925107, 62271281, 62021002)
%
%
\bibliographystyle{splncs04}
\bibliography{egbib}

\appendix
\section{Datasets}

Building upon prior research~\cite{zhou2021coop,zhou2022conditional}, we utilize eleven datasets pertaining to image recognition to substantiate the effectiveness of the proposed methodology in addressing the base-to-new generalization task. These datasets encompass two repositories dedicated to generic object classification, namely ImageNet and Caltech101, five repositories catering to fine-grained classification, including OxfordPets, StanfordCars, Flowers102, Food101, and FGVCAircraft, one repository for scene recognition, denoted as SUN397, one repository for action recognition, known as UCF101, one repository for texture classification, termed DTD, and one repository for satellite imagery recognition, designated EuroSAT. Consistent with earlier studies~\cite{zhou2021coop,zhou2022conditional,khattak2023maple,zhu2023prograd}, for each dataset in base-to-new generalization, we evenly partition the classes into two distinct groups that do not overlap, with one group serving as the base classes and the other as the new classes. We train all models only using the base classes and conduct evaluation on both the base and new classes to verify the specialization capability and generalization capability of the models.

In the domain generalization task, we leverage ImageNet-A, ImageNet-R, ImageNetv2, and ImageNet-S to assess the robustness of the model. In this context, the model is initially trained using ImageNet, followed by direct utilization of images from the aforementioned datasets for inference.

Concerning the cross-dataset transfer task, the datasets mirror those utilized in the base-to-new generalization task. Analogous to domain generalization, the model undergoes initial training on ImageNet followed by inference on the remaining ten distinct datasets.

For the few-shot learning task, the datasets align with those employed in the base-to-new generalization task. The model is trained and assessed with varying numbers of shots, specifically 1, 2, 4, 8, and 16 shots separately.

The dataset partitioning mirrors that of earlier works~\cite{zhou2021coop,zhou2022conditional}. We present the average model performance over three iterations with distinct random seeds to ensure fair comparisons.

\begin{table}[htbp]
    \tiny
    \setlength{\belowcaptionskip}{-0.05cm}
    \setlength{\tabcolsep}{1.2mm}
    \caption{Full results in base-to-new generalization. H: harmonic mean~\cite{xian2017zero}.}
    \label{tab:appendix_results_base_to_new_generalization}
    \begin{subtable}[t]{\textwidth}
    \centering
    \caption{\textbf{An overview of the size of different methods.}.}
    \begin{tabular}{l|cccccccc}
    \toprule
    Method & CoOp & CoCoOp & Adapter & LoRA & ProGrad & QCoOp & MaPLe & QMaPLe\\
    \midrule
    size & 4.1KB & 70.8KB & 1051KB & 258KB & 16.4KB & 0.26KB & 7096KB & 1774KB\\
    \bottomrule
    \end{tabular}
    \end{subtable}
    \begin{subtable}[t]{.3\textwidth}
    \centering
    \caption{Average}
    \begin{tabular}{lcc|c}
    \toprule
    & Base & New & H \\
    \midrule
    CLIP & 69.34 & 74.22 & 71.70 \\
    CoOp  & 82.69 & 63.22 & 71.66 \\
    CoCoOp & 80.47 & 71.69 & 75.83 \\
    Adapter & 82.62 & 70.97 & 76.35 \\
    LoRA & 84.30 & 67.33 & 74.86 \\
    ProGrad & 82.79 & 68.55 & 75.00 \\
    \rowcolor{tabhighlight}
    QCoOp & 80.68 & 74.44 & 77.43 \\
    \hline
    MaPLe & 82.28 & 75.14 & 78.55 \\
    \rowcolor{tabhighlight}
    QMaPLe & 83.02 & 75.57 & 79.12 \\
    \bottomrule
    \end{tabular}
    \end{subtable}
    \hspace{1em}
    \vspace{1em}
    \begin{subtable}[t]{.3\textwidth}
    \centering
    \caption{ImageNet}
    \begin{tabular}{lcc|c}
    \toprule
    & Base & New & H \\
    \midrule
    CLIP & 72.43 & 68.14 & 70.22 \\
    CoOp & 76.47 & 67.88 & 71.92\\
    CoCoOp & 75.98 & 70.43 & 73.10 \\
    Adapter & 76.53 & 66.67 & 71.26 \\
    LoRA & 74.77 & 58.47 & 65.62\\
    ProGrad & 77.03 & 68.80 & 72.68 \\
    \rowcolor{tabhighlight}
    QCoOp & 76.17 & 70.73 & 73.35 \\
    \hline
    MaPLe & 76.66 & 70.54 & 73.47 \\
    \rowcolor{tabhighlight}
    QMaPLe & 76.93 & 70.73 & 73.70 \\
    \bottomrule
    \end{tabular}
    \end{subtable}
    ~
    \begin{subtable}[t]{.3\textwidth}
    \centering
    \caption{Caltech101}
    \begin{tabular}{l cc|c}
    \toprule
    & Base & New & H \\
    \midrule
    CLIP & 96.84 & 94.00 & 95.40 \\
    CoOp & 98.00 & 89.81 & 93.73 \\
    CoCoOp & 97.96 & 93.81 & 95.84 \\
    Adapter & 98.20 & 93.20 & 95.63 \\
    LoRA & 98.49 & 90.33 & 94.24\\
    ProGrad & 98.50 & 91.90 & 95.09 \\
    \rowcolor{tabhighlight}
    QCoOp & 97.80 & 95.03 & 96.40 \\ 
    \hline
    MaPLe & 97.74 & 94.36 & 96.02 \\
    \rowcolor{tabhighlight}
    QMaPLe & 97.97 & 95.00 & 96.46 \\
    \bottomrule
    \end{tabular}
    \end{subtable}
    ~
    \begin{subtable}[t]{.3\textwidth}
    \centering
    \caption{OxfordPets}
    \begin{tabular}{l cc|c}
    \toprule
    & Base & New & H \\
    \midrule
    CLIP & 91.17 & 97.26 & 94.12 \\
    CoOp & 93.67 & 95.29 & 94.47 \\
    CoCoOp & 95.20 & 97.69 & 96.43 \\
    Adapter & 94.40 & 94.10 & 94.25 \\
    LoRA & 94.90 & 92.57 & 93.72\\
    ProGrad & 94.40 & 95.10 & 94.75 \\
    \rowcolor{tabhighlight}
    QCoOp & 95.17 & 97.60 & 96.37\\
    \hline
    MaPLe & 95.43 & 97.76 & 96.58 \\
    \rowcolor{tabhighlight}
    QMaPLe & 95.67 & 97.63 & 96.64 \\
    \bottomrule
    \end{tabular}
    \end{subtable}
    \hspace{1em}
    \vspace{1em}
    \begin{subtable}[t]{.3\textwidth}
    \centering
    \caption{StanfordCars}
    \begin{tabular}{l cc|c}
    \toprule
    & Base & New & H \\
    \midrule
    CLIP & 63.37 & 74.89 & 68.65 \\
    CoOp & 78.12 & 60.40 & 68.13 \\
    CoCoOp & 70.49 & 73.59 & 72.01 \\
    Adapter & 77.13 & 69.23 & 72.97 \\
    LoRA & 81.07 & 65.30 & 72.34 \\
    ProGrad & 79.00 & 67.93 & 73.05 \\
    \rowcolor{tabhighlight}
    QCoOp & 73.73  & 72.90 & 73.31 \\
    \hline
    MaPLe & 72.94 & 74.00 & 73.47 \\
    \rowcolor{tabhighlight}
    QMaPLe & 75.00 & 73.67 & 74.33 \\
    \bottomrule
    \end{tabular}
    \end{subtable}
    ~
    \begin{subtable}[t]{.3\textwidth}
    \centering
    \caption{Flowers102}
    \begin{tabular}{l cc|c}
    \toprule
    & Base & New & H \\
    \midrule
    CLIP & 72.08 & 77.80 & 74.83 \\
    CoOp & 97.60 & 59.67 & 74.06 \\
    CoCoOp & 94.87 & 71.75 & 81.71 \\
    Adapter & 97.70 & 70.83 & 82.13 \\
    LoRA & 98.23 & 60.20 & 74.65\\
    ProGrad & 96.27 & 71.07 & 81.77 \\
    \rowcolor{tabhighlight}
    QCoOp & 95.57 & 74.67 & 84.22\\
    \hline
    MaPLe & 95.92 & 72.46 & 82.56 \\
    \rowcolor{tabhighlight}
    QMaPLe & 96.43 & 74.33 & 83.95 \\
    \bottomrule
    \end{tabular}
    \end{subtable}
    ~
    \begin{subtable}[t]{.3\textwidth}
    \centering
    \caption{Food101}
    \begin{tabular}{l cc|c}
    \toprule
    & Base & New & H \\
    \midrule
    CLIP & 90.10 & 91.22 & 90.66 \\
    CoOp & 88.33 & 82.26 & 85.19 \\
    CoCoOp & 90.70 & 91.29 & 90.99 \\
    Adapter & 90.40 & 90.40 & 90.40 \\
    LoRA & 88.57 & 87.30 & 87.93\\
    ProGrad & 90.17 & 89.53 & 89.85 \\
    \rowcolor{tabhighlight}
    QCoOp & 90.87  & 91.90 & 91.38\\
    \hline
    MaPLe & 90.71 & 92.05 & 91.38 \\
    \rowcolor{tabhighlight}
    QMaPLe & 90.63 & 92.10 & 91.36 \\
    \bottomrule
    \end{tabular}
    \end{subtable}
    \hspace{1em}
    \vspace{1em}
    \begin{subtable}[t]{.3\textwidth}
    \centering
    \caption{FGVCAircraft}
    \begin{tabular}{l cc|c}
    \toprule
    & Base & New & H \\
    \midrule
    CLIP & 27.19 & 36.29 & 31.09 \\
    CoOp & 40.44 & 22.30 & 28.75 \\
    CoCoOp & 33.41 & 23.71 & 27.74 \\
    Adapter & 39.57 & 32.27 & 35.55 \\
    LoRA & 46.27 & 28.83 & 35.53\\
    ProGrad & 42.63 & 26.97 & 33.04 \\
    \rowcolor{tabhighlight}
    QCoOp & 37.50 & 34.03 & 35.68  \\
    \hline
    MaPLe & 37.44 & 35.61 & 36.50 \\
    \rowcolor{tabhighlight}
    QMaPLe & 39.10 & 34.90 & 36.88 \\
    \bottomrule
    \end{tabular}
    \end{subtable}
    ~
    \begin{subtable}[t]{.3\textwidth}
    \centering
    \caption{SUN397}
    \begin{tabular}{l cc|c}
    \toprule
    & Base & New & H \\
    \midrule
    CLIP & 69.36 & 75.35 & 72.23 \\
    CoOp & 80.60 & 65.89 & 72.51 \\
    CoCoOp & 79.74 & 76.86 & 78.27 \\
    Adapter & 81.67 & 73.93 & 77.61 \\
    LoRA & 79.73 & 69.00 & 73.98\\
    ProGrad & 80.70 & 71.03 & 75.56 \\
    \rowcolor{tabhighlight}
    QCoOp & 79.20 & 77.93 & 78.56 \\
    \hline
    MaPLe & 80.82 & 78.70 & 79.75 \\
    \rowcolor{tabhighlight}
    QMaPLe & 81.33 & 78.27 & 79.77 \\
    \bottomrule
    \end{tabular}
    \end{subtable}
    ~
    \begin{subtable}[t]{.3\textwidth}
    \centering
    \caption{DTD}
    \begin{tabular}{lcc|c}
    \toprule
    & Base & New & H \\
    \midrule
    CLIP & 53.24 & 59.90 & 56.37 \\
    CoOp & 79.44 & 41.18 & 54.24 \\
    CoCoOp & 77.01 & 56.00 & 64.85 \\
    Adapter & 80.47 & 52.23 & 63.35 \\
    LoRA & 82.93 & 54.90 & 66.06\\
    ProGrad & 76.70 & 46.67 & 58.03 \\
    \rowcolor{tabhighlight}
    QCoOp & 74.97 & 58.37 & 65.63 \\
    \hline
    MaPLe & 80.36 & 59.18 & 68.16 \\
    \rowcolor{tabhighlight}
    QMaPLe & 80.77 & 57.63 & 67.27 \\
    \bottomrule
    \end{tabular}
    \end{subtable}
    \hspace{2.1em}
    \begin{subtable}[t]{.3\textwidth}
    \centering
    \caption{EuroSAT}
    \begin{tabular}{lcc|c}
    \toprule
    & Base & New & H \\
    \midrule
    CLIP & 56.48 & 64.05 & 60.03 \\
    CoOp & 92.19 & 54.74 & 68.69 \\
    CoCoOp & 87.49 & 60.04 & 71.21 \\
    Adapter & 86.93 & 64.20 & 73.86 \\
    LoRA & 94.90 & 65.67 & 77.62\\
    ProGrad & 91.37 & 56.53 & 69.85 \\
    \rowcolor{tabhighlight}
    QCoOp & 83.53 & 69.80 & 76.05 \\
    \hline
    MaPLe & 94.07 & 73.23 & 82.35 \\
    \rowcolor{tabhighlight}
    QMaPLe & 94.30 & 79.47 & 86.25 \\
    \bottomrule
    \end{tabular}
    \end{subtable}
    \hspace{2em}
    \begin{subtable}[t]{.3\textwidth}
    \centering
    \caption{UCF101}
    \begin{tabular}{lcc|c}
    \toprule
    & Base & New & H \\
    \midrule
    CLIP & 70.53 & 77.50 & 73.85 \\
    CoOp & 84.69 & 56.05 & 67.46 \\
    CoCoOp & 82.33 & 73.45 & 77.64 \\
    Adapter & 85.80 & 73.63 & 79.25 \\
    LoRA & 87.47 & 68.03 & 76.53\\
    ProGrad & 83.90 & 68.50 & 75.42 \\
    \rowcolor{tabhighlight}
    QCoOp & 81.87 & 75.93 & 78.79 \\
    \hline
    MaPLe & 83.00 & 78.66 & 80.77 \\
    \rowcolor{tabhighlight}
    QMaPLe & 85.10 & 77.50 & 81.12 \\
    \bottomrule
    \end{tabular}
    \end{subtable}
\end{table}

\section{Training Configuration} 
Following the conventional setup outlined in~\cite{zhou2022conditional}, we employ ViT-B/16 as the image encoder within CLIP. Prior to feeding into the image encoder, each training image is resized to $224\times224$. To augment the data, standard techniques such as random cropping and flipping are applied, consistent with the methodology described in~\cite{zhou2022conditional}. During training, a batch size of 32 is utilized, and stochastic gradient descent (SGD) is employed to optimize the learnable parameters. Similar to the approach detailed in~\cite{zhou2021coop}, a warm-up scheme is implemented during the first epoch, which proves crucial for prompt tuning. All other baselines are configured strictly according to the specifications provided in their respective original papers.

Hyperparameter tuning is performed via a grid search methodology, guided by the parameter configurations reported in previous studies~\cite{zhou2021coop,khattak2023maple}. For experiments involving QCoOp, the quantization bit of the prompts is set to 1 by default unless otherwise specified. In the case of QMaPLe experiments, the parameters in the projection layer, responsible for transforming text prompts into image prompts, and the parameters in the prompts are all subjected to quantization, with a quantization bit of 4. Additionally, following the structure of MaPLe, nine transformer layers are typically modified within QMaPLe experiments.

\begin{table}[tb]
    \centering
    \scriptsize
    \setlength{\belowcaptionskip}{-0.05cm}
    \caption{Results of deep prompts for QCoOp.}
    \setlength{\tabcolsep}{2.7mm}{
    \renewcommand{\arraystretch}{1.2}
    \begin{tabular}{|c|c|c|c|c|}
        \hline
        total depth & size & base & new & H \\
        \hline
        1 & 0.26KB & 79.49 & 72.65 & 75.92 \\
        \hline
        2 & 0.52KB & 78.19 & 73.67 & 75.86\\
        \hline
        3 & 0.78KB & 79.82 & 72.27 & 75.86 \\
        \hline
        4 & 1.04KB & 80.77 & 72.31 & 76.31\\
        \hline
        5 & 1.30KB & 81.14 & 72.67 & 76.67 \\
        \hline
        6 & 1.56KB & 81.65 & 72.68 & 76.90\\
        \hline
    \end{tabular}
    }
    \label{tab:deepprompt}
\end{table}

\section{Full Base-To-New Generalization Results}
As shown in \cref{tab:appendix_results_base_to_new_generalization}, QCoOp and QMaPLe significantly improve the generalization capability represented by the accuracy on the new classes. Compared with CoOp, QCoOp earns $11.22$\% accuracy gain on the new classes and $2.10$\% accuracy drop on the base classes. Among all the lightweight SOTA methods, QCoOp gets strongest harmonic mean accuracy on 7 out of 11 datasets including ImageNet, Caltech101, StanfordCars, Flowers102, Food101, FGVCAircraft, and SUN397. Compared with a heavy method MaPLe, QMaPLe achieves better harmonic mean accuracy on 9 out of 11 datasets, including ImageNet, Caltech101, OxfordPets, StanfordCars, Flowers102, FGVCAircraft, SUN397, EuroSAT and UCF101.

\section{Experiments on Deep Prompt Configurations}
In this paragraph, we investigate the impact of deep prompts, as described in Eq. 4 of the main text. By default, QCoOp employs prompts with a depth of 1, which means the prompts are only added to the input layer. We systematically increase the number of tunable prompts in subsequent transformer layers. Considering that the text transformer in CLIP consists of 12 layers, we add prompts to a maximum of 6 layers for verification. As illustrated in Table \cref{tab:deepprompt}, incorporating more prompts across multiple layers tends to improve the specialized capability, albeit at the expense of a slight reduction in generalized capability. In summary, at the expense of increased dimensionality, using more prompts across multiple layers can improve the harmonic mean accuracy to some extent.

\begin{table}[tb]
    \centering
    \scriptsize
    \setlength{\belowcaptionskip}{-0.05cm}
    \caption{Comparisons of using quantized prompts in textual encoder and image encoder at minimal size.}
    \setlength{\tabcolsep}{2.7mm}{
    \renewcommand{\arraystretch}{1.2}
    \begin{tabular}{|c|c|c|c|c|}
        \hline
         & size & base & new & H \\
        \hline
        QCoOp & 0.26KB & 79.49 & 72.65 & 75.92 \\
        \hline
        QVPT & 0.39KB & 73.15 & 68.93 & 70.98\\
        \hline
    \end{tabular}
    }
    \label{tab:diff_modal}
\end{table}

\section{Prompt Modality Considerations for Minimizing Storage Cost}
In this paragraph, we conduct a brief comparison of quantizing prompts in the visual transformer, following the approach outlined in VPT~\cite{jia2022vpt}. The results are presented in \cref{tab:diff_modal}. It is evident that when operating under a stringent storage constraint, QCoOp outperforms QVPT in terms of both accuracy and size efficiency. Adding prompts to the textual transformer is better.

\end{document}